\documentclass[11pt]{article}

\usepackage[preprint]{acl}

\usepackage{times}
\usepackage{latexsym}

\usepackage{amsmath}
\usepackage{booktabs}   
\usepackage{caption}    
\usepackage{amssymb}

\usepackage[T1]{fontenc}

\usepackage[utf8]{inputenc}

\usepackage{microtype}

\usepackage{inconsolata}

\usepackage{graphicx}
\usepackage{subcaption} 

\title{Scaling Probabilistic Transformer via Efficient Cross-Scale Hyperparameter Transfer}

\author{
  Penghao Kuang,
  Haoyi Wu,
  Kewei Tu\textsuperscript{*}
  \\
  School of Information Science and Technology, ShanghaiTech University\\
  Shanghai Engineering Research Center of Intelligent Vision and Imaging\\
  \begingroup
    \fontfamily{lmr}\selectfont 
    \{kuangph2023,wuhy1,tukw\}@shanghaitech.edu.cn
  \endgroup\\
}

\begin{document}
\maketitle
\begin{abstract}
Probabilistic Transformer (PT), a white-box probabilistic model for contextual word representation, has demonstrated substantial similarity to standard Transformers in both computational structure and downstream task performance on small models and small to medium sized datasets. However, PT is less robust to hyperparameter choices than standard Transformers, making it harder to scale efficiently. In this work, we follow Maximal Update Parametrization ($\mu$P) to rescale PT’s parameters, so that hyperparameters optimized on small models can be transferred to larger models without additional tuning. With this approach, we successfully scale PT to models with up to 0.4B parameters. Experiments show that PT consistently outperforms standard transformer under the same parameter budget on Masked Language Modeling (MLM) tasks. We hope this work will contribute to the practical deployment of probabilistic models at substantially larger scales in the future.
\end{abstract}

\section{Introduction}

\begin{table*}[t]
\centering
\caption{Grouping of learnable parameters}
\label{tab:grouping}
\begin{tabular}{@{}ll@{}}
\toprule
\textbf{Category} & \textbf{Learnable Parameters} \\
\midrule
\textbf{Input Weights} & 
\begin{tabular}[c]{@{}l@{}}Unary factor ($S$ in PT) \\ LayerNorm Weight of MLM Head ($\gamma$ of RMSNorm) \\ All Biases\end{tabular} \\
\midrule
\textbf{Hidden Weights} & 
\begin{tabular}[c]{@{}l@{}}Ternary Factor $U$ ($U$ in PT)\\ Ternary Factor $V$ ($V$ in PT)\\ Binary Factor ($B$ in PT)\end{tabular} \\
\midrule
\textbf{Output Weights} & Decoder Weight of MLM Head \\
\bottomrule
\end{tabular}
\end{table*}

Neural networks, represented by the Transformer architecture \cite{vaswani2017attention}, have achieved tremendous success in fields such as natural language processing. However, as data-driven "black-box" models, they lack intrinsic mathematical transparency and theoretical guarantees. To overcome this bottleneck, "white-box models" have gradually become a frontier focus in deep learning architecture design. For instance, the Probabilistic Transformer (PT) \cite{wu2023probabilistic} adopts a purely syntactic and probabilistic perspective, introducing Conditional Random Fields (CRFs) \cite{lafferty2001conditional} to model latent representations and employing Mean Field Variational Inference (MFVI) \cite{blei2017variational} for approximate inference. This intrinsically interpretable architecture not only demonstrates performance approaching that of the standard Transformer on small model and small- to medium-scale datasets, but also provides crucial insights for revealing the underlying mechanisms of the model.

However, PT faces a severe bottleneck in scaling its model parameters. When executing variational inference modules such as Head Selection and Topic Modeling \cite{blei2003latent}, the PT requires numerical scaling of the message tensors prior to non-linear normalization (e.g., Softmax), thereby introducing 6 additional information weight hyperparameters that do not exist in standard Transformers. These 6 hyperparameters exhibit highly non-linear coupling with the learning rate, rendering traditional single-variable tuning strategies completely ineffective. Furthermore, their optimal configurations experience drastic shifts as the model scale increases. Under constrained computational resources, this complex hyperparameter space, which lacks cross-scale transferability, makes hyperparameter search at large model scales prohibitively expensive or even infeasible, severely restricting the potential of the PT to evolve toward large-scale parameters.

Maximal Update Parametrization ($\mu$P) \cite{yang2022tensor} is an effective method for resolving hyperparameter drift with scale, enabling the zero-shot transfer of optimal hyperparameters to large models. However, the current $\mu$P framework is specifically designed for "black-box" networks, and its core operations involve scaling adjustments to intermediate activations or logits (for example, to ensure variance stability, the logits scaling factor in the attention mechanism is modified from $1/\sqrt{d_k}$ to $1/d_k$). In the computation graph of the PT, the activation values at each layer represent strict probability distributions. Directly applying scaling factors to these values violates the probability distribution assumptions, thereby destroying the white-box nature of the model.

To address the aforementioned theoretical conflict, this paper proposes a cross-scale parameterization reconstruction method tailored for the PT. Drawing upon the underlying principles of $\mu$P, we systematically apply scaling adjustments to the potential functions and the variational free energy of the PT, under the strict premise of maintaining the probability distribution assumptions. Concurrently, we modify the parameter initialization variance and the group-specific learning rates for particular modules. The adjusted PT not only strictly preserves the interpretable mathematical semantics of the activations at each layer, but also attains a hyperparameter transfer capability consistent with that of black-box networks utilizing $\mu$P. This method enables the PT to be optimally tuned directly on extremely small-scale models and seamlessly transferred to any parameter scale, substantially reducing scaling costs. Experiments demonstrate that across Masked Language Modeling (MLM) tasks of varying scales, the PT architecture equipped with this scaling capability consistently outperforms BERT \cite{devlin2019bert} and achieves performance approaching that of the Universal Transformer \cite{dehghani2019universal}.

\section{Adjustment of $\mu$P to PT}

This section aims to elaborate on the cross-scale scaling framework tailored for the Probabilistic Transformer (PT). By systematically reconstructing the parameterization and underlying mathematical architecture, we successfully enable the PT to achieve zero-shot transferability for global hyperparameters—including the learning rate and 6 core information weights—under the strict premise of preserving the white-box property of its probabilistic inference. Theoretical analysis indicates that our architectural adjustments are mathematically consistent with the infinite-width scaling principles of $\mu$P. Rigorous derivations and proofs of the relevant theorems are detailed in Appendix~\ref{app:C}.

\subsection{Adjustment of parameterization}

\textbf{Grouping of Learnable Parameters.} $\mu$P categorizes learnable parameters based on the relationship between parameter size and model width. In the PT architecture, the dimension of the latent variable $Z$ nodes (denoted as $\text{dim}_z$) naturally corresponds to the model width. Based on the classification criteria of $\mu$P, we strictly partition all learnable parameters in the PT into three categories: Input Weights, Hidden Weights, and Output Weights. The specific parameter mapping is detailed in Table~\ref{tab:grouping}.\\

\noindent
\textbf{Reconstruction of Initialization Distributions and Group-specific Learning Rates.} Following the theoretical derivations of $\mu$P, we tailor the initialization variance scaling rules and group-specific learning rates for different parameter categories. Specifically, the hidden and output weights are initialized with variances scaled by the model width $N$ (i.e., $\text{dim}_z$) and are assigned corresponding scaled learning rates to ensure the stability of training dynamics. Meanwhile, the input weights and biases maintain the base distribution and base learning rate. The detailed cross-scale parameterization configuration is presented in Table~\ref{tab:init_lr}.

\begin{table*}[t]
\centering
\caption{Initialization distribution of learnable parameters and learning rate reconstruction}
\label{tab:init_lr}
\begin{tabular}{@{}lll@{}}
\toprule
\textbf{Parameter Group} & \textbf{Initialization Distribution} & \textbf{Learning Rate} \\
\midrule
\textbf{Input Weights} (excluding Biases) & $\mathcal{N}(0, 1)$ & \textbf{Base LR ($\eta$)} \\
\midrule
\textbf{Hidden Weights} & $\mathcal{N}\left(0, \frac{1}{N}\right)$ & \textbf{Scaled LR ($\frac{\eta}{N}$)} \\
\midrule
\textbf{Output Weights} & $\mathcal{N}\left(0, \frac{1}{N^2}\right)$ & \textbf{Scaled LR ($\frac{\eta}{N}$)} \\
\midrule
\textbf{Biases} & $0$ & \textbf{Base LR ($\eta$)} \\
\bottomrule
\end{tabular}
\end{table*}

\subsection{Adjustment of mathematical architecture}

\textbf{Scaling Paradigms for Head Selection.} In the Head Selection module of the PT, we construct a dependency matrix $T\in \mathbb{R}^{N\times N}$ comprising $C$ channels. For the matrix $T$ of each channel, we employ low-rank factorization to express it as $UV^T$, where $U,V\in \mathbb{R}^{N\times r}$. As the model width $N$ expands, the parameterization scaling of the Head Selection module faces two viable design paradigms: the first fixes the rank $r$ while the number of channels $C$ scales linearly with $N$; the second fixes $C$ while $r$ scales linearly with $N$. To ensure the universality of the architecture, the reconstruction of the PT's underlying mathematical architecture must simultaneously accommodate both scaling schemes, thereby enabling a fair evaluation to select the optimal scaling strategy.\\

\noindent
\textbf{Reconstruction of Potential Functions and Variational Free Energy.} In the probabilistic graphical model of the PT, the core inference process relies on executing Mean Field Variational Inference (MFVI) for the posterior probability distribution $Q_i$ of the latent variable $Z$ nodes and the posterior probability distribution $Q_{ic}$ of the $H$ nodes. To introduce cross-scale numerical control, we define a temperature parameter for the $Q_i$ distribution:
\begin{equation}
\tau=\frac{N}{r}
\end{equation}
while the temperature parameter for the $Q_{ic}$ distribution is held constant at 1. Furthermore, we define the number of $G$ nodes as $M$, which scales proportionally with $N$. Building upon this foundation, and in contrast to the original PT architecture, we reconstruct the potential functions of the system as follows:
\begin{equation}
\phi_u(Z_i=a)=\exp(\tau S_{w_i,a}) 
\end{equation}
\begin{equation}
\begin{split}
\phi_t(H_i=k,Z_i=a,Z_j=b) = & \\
\begin{cases} 
    \exp(\tau N T_{a,b}) & H_i=j \\ 
    1 & \text{otherwise} 
\end{cases} &
\end{split}
\end{equation}
\begin{equation}
\phi_b(Z_i=a,G_i=g)=\exp(\tau M B_{g,a})
\end{equation}
Correspondingly, the variational free energy function for $Q_i$ is modified as follows:
\begin{equation}
F(Q_i)=\mathbb{E}_Q[-\log P(Z,H,G|w)]-\tau H(Q_i)
\end{equation}
while the variational free energy function for $Q_{ic}$ remains unchanged:
\begin{equation}
F(Q_{ic})=\mathbb{E}_Q[-\log P(Z,H,G|w)]-H(Q_{ic})
\end{equation}
regardless of whether "scaling the number of channels $C$" or "scaling the rank $r$" is chosen, the Head Selection module can achieve rigorous $\mu$P scaling within this unified mathematical framework. We provide detailed mathematical proofs in Appendix~\ref{app:B}.\\

\noindent
\textbf{Mathematical Equivalence to the Black-box $\mu$P Mechanism.} Unlike standard black-box neural networks, which can apply scalar multipliers directly to activations for scaling, the PT must maintain the probability distribution assumptions of its activations. Therefore, by modifying the potential functions and variational free energy, we achieve scaling while preserving the white-box nature of the PT. It can be proven that the aforementioned adjustments in the analytical derivation of the final computation graph are mathematically equivalent to the direct numerical scaling used in black-box models. The effect is strictly equivalent to scaling two key activations during the MFVI iteration:
\begin{equation}
\begin{cases}
\mathcal{F}_c^{(t)} \leftarrow \mathcal{F}_c^{(t)}/r \\
Q_z^{(t)} \leftarrow Q_z^{(t)} \cdot N
\end{cases}
\end{equation}
It is worth emphasizing that in the inference logic of the PT, the function of the message tensor $\mathcal{F}_c^{(t)}$ completely corresponds to the attention logits in standard Transformers. Thus, implicitly scaling it as $\mathcal{F}_c^{(t)}/r$ mathematically aligns with the core operation in $\mu$P theory of modifying the attention scaling factor to $qk^T/d_k$. Rigorous derivations of this equivalence property are provided in Appendix~\ref{app:A}.

\begin{figure*}[t] 
    \centering
    
    \begin{subfigure}[t]{0.495\textwidth} 
        \centering
        \includegraphics[width=\linewidth]{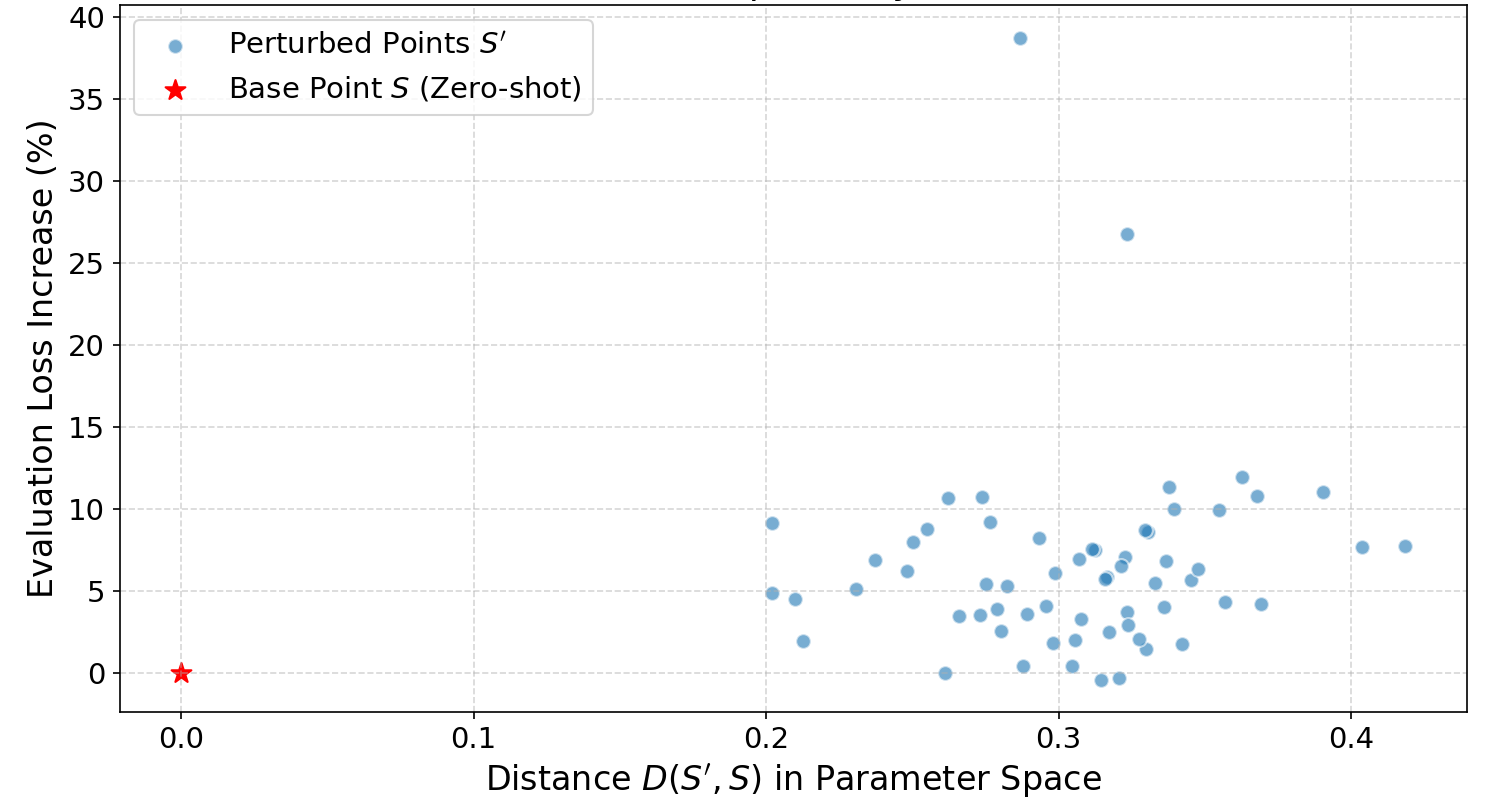} 
        \caption{Distance vs Evaluation Loss Increase.}
        \label{fig:left_plot1}
    \end{subfigure}
    \hfill
    \begin{subfigure}[t]{0.495\textwidth}
        \centering
        \includegraphics[width=\linewidth]{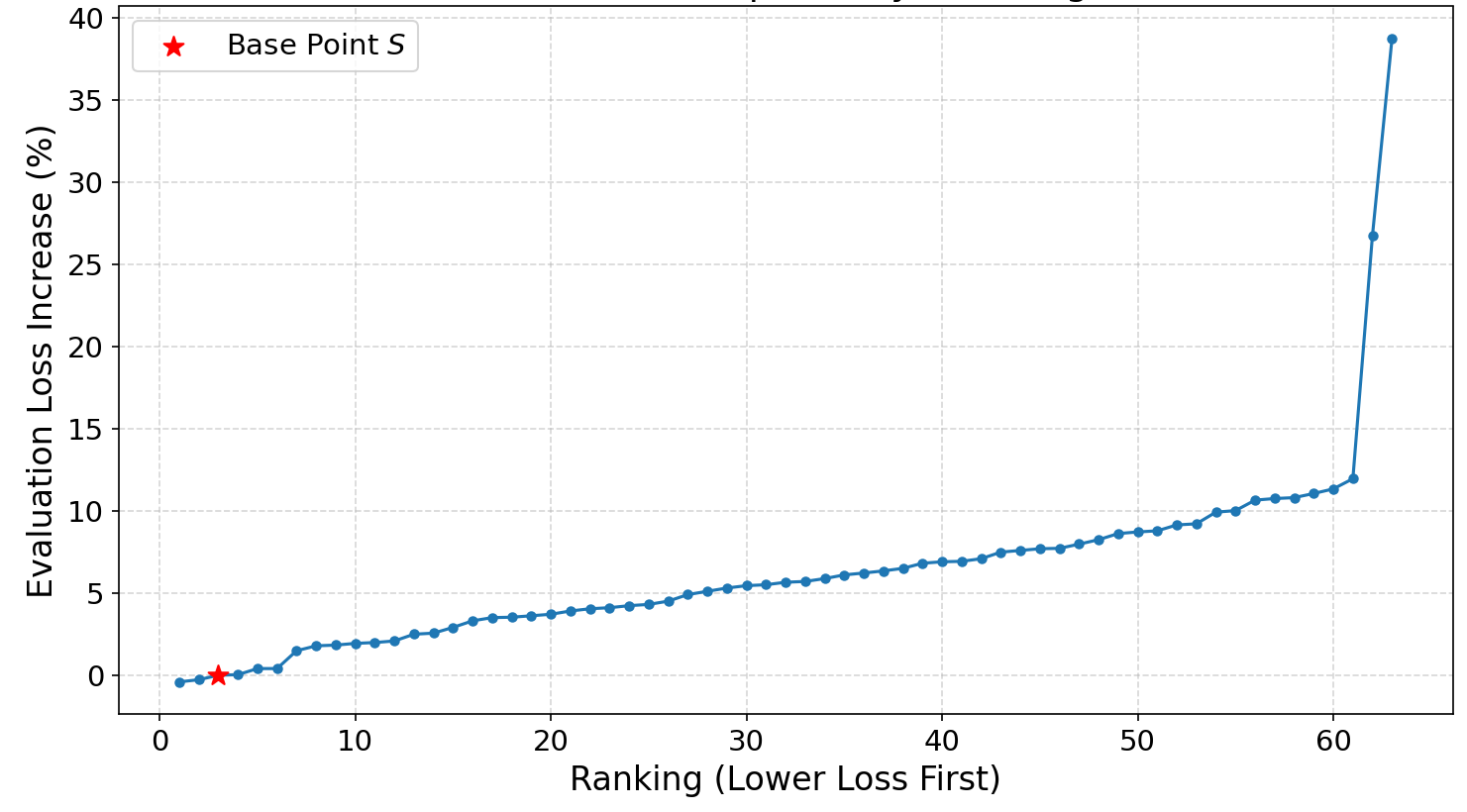}
        \caption{Rank vs Evaluation Loss Increase.}
        \label{fig:right_plot1}
    \end{subfigure}
    
    \caption{
       Performance comparison between perturbed sampling points and benchmark points.
    }
    \label{fig:combined_charts}
\end{figure*}

\section{Experiments}

\subsection{Optimality Verification of Scaling}

\textbf{Random Search Verification.} Given the constraints of computational resources, executing a global grid search within a highly coupled 7-dimensional hyperparameter space to verify the optimality of the zero-shot transferred configuration is computationally infeasible. Therefore, we employ a local optimality verification method based on random search \cite{bergstra2012random}. Specifically, by independently sampling $n$ configurations within a local neighborhood of the zero-shot transferred hyperparameters for evaluation, if the evaluation loss of the original configuration remains within the minimal range among the perturbed samples, we can statistically prove, with a confidence of $1-\alpha$, that the original configuration belongs to the top $p$ fraction of the optimal parameter subspace in that neighborhood. This confidence level satisfies the following inequality constraint:
\begin{equation}
1-(1-p)^n \geq 1-\alpha
\end{equation}
Given the target superiority proportion $p$ and the significance level $\alpha$, the minimum required number of independent samples is:
\begin{equation}
n \geq \frac{\log \alpha}{\log (1-p)}
\end{equation}
\\

\noindent
\textbf{Methods.} This experiment is conducted on a PT model with a width of $\text{dim}_z=1536$. Let the hyperparameter set obtained via zero-shot transfer be the central base point $S=\{S_1, S_2, \cdots, S_7\}$. For any $S_i\ (1 \leq i \leq 7)$, we perform independent random sampling within a uniform distribution interval of $\pm 20\%$, i.e., $[0.8S_i, 1.2S_i]$, thereby constructing a perturbed sampling set $S'$ within the local neighborhood of the base point $S$. To verify with 95\% confidence (i.e., $\alpha=0.05$) whether $S$ falls into the top 5\% (i.e., $p=0.05$) optimal region within the local space, the theoretical minimum number of samples is calculated as $n \geq 58.4$ according to the aforementioned theorem. To this end, we execute $n=62$ independent local random sampling training runs and quantitatively compare their final evaluation losses with that of the base point $S$.\\

\noindent
\textbf{Results.} The experimental results exhibit a highly significant basin of local minimum effect. Among all 62 perturbed samples, the evaluation losses of the vast majority of configurations increased compared to the zero-shot configuration. Although a small number of optimal sampling points exist, their loss exhibited a marginal decrease of at most 0.4\% relative to $S$. Considering the stochastic noise inherent in the optimization process, such extremely subtle performance fluctuations fall perfectly within the training tolerance range and can be deemed statistically equivalent to $S$. To intuitively quantify this phenomenon, we define the distance between the perturbed sampling point $S'$ and the base point $S$ in the relative parameter space as:
\begin{equation}
D(S',S)=\sqrt{\sum_{i=1}^7 \left(\frac{S_i'-S_i}{S_i}\right)^2}
\end{equation}
As illustrated in Figure~\ref{fig:combined_charts}, we visualize the evaluation results from two perspectives: parameter space distance and relative performance ranking. The distance scatter plot demonstrates the performance degradation trend associated with deviations from the base point, while the full-sample loss ranking curve corroborates that $S$ (marked with a red star) securely resides at the very bottom of this steep local minimum basin.  Therefore, statistically, we possess 95\% confidence to assert that the hyperparameter configuration acquired through zero-shot transfer successfully locates within the top 5\% optimal region of this neighborhood space.

\begin{figure*}[t] 
    \centering
    
    \begin{subfigure}[t]{0.495\textwidth} 
        \centering
        \includegraphics[width=\linewidth]{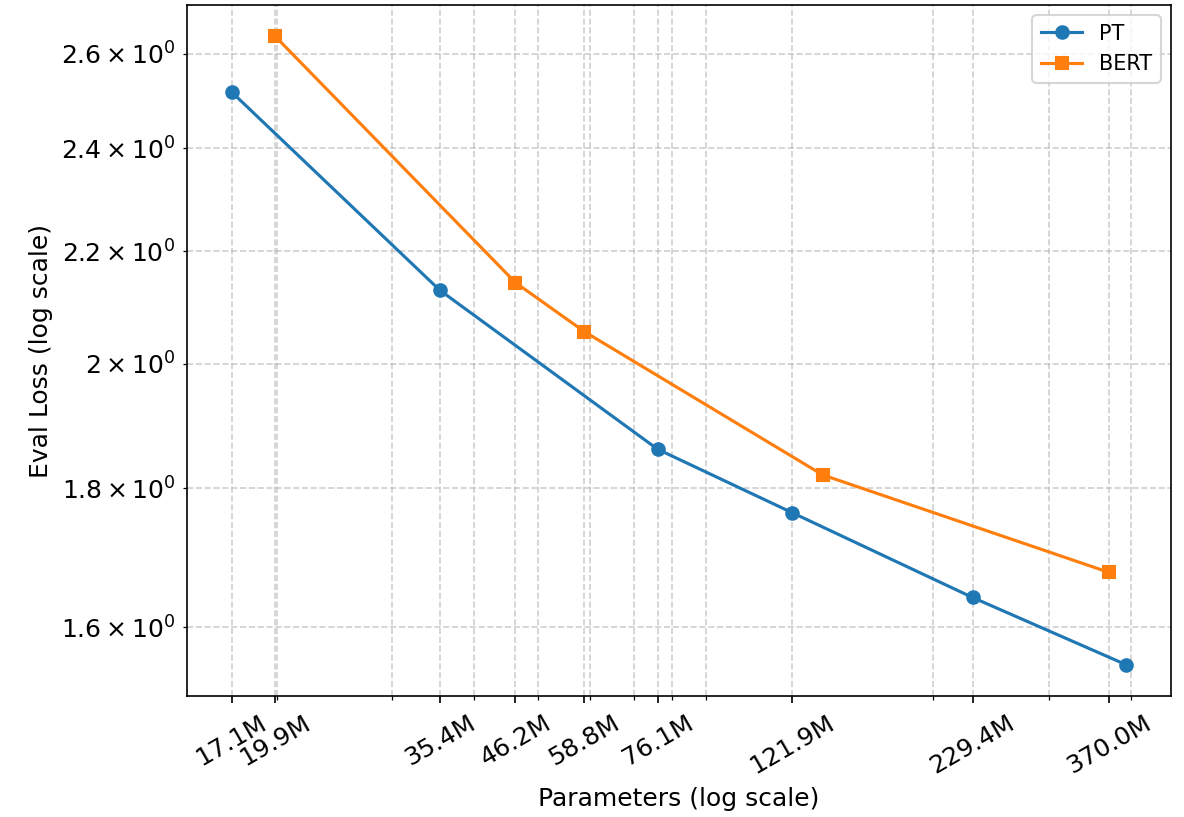} 
        \caption{PT vs BERT.}
        \label{fig:left_plot2}
    \end{subfigure}
    \hfill
    \begin{subfigure}[t]{0.495\textwidth}
        \centering
        \includegraphics[width=\linewidth]{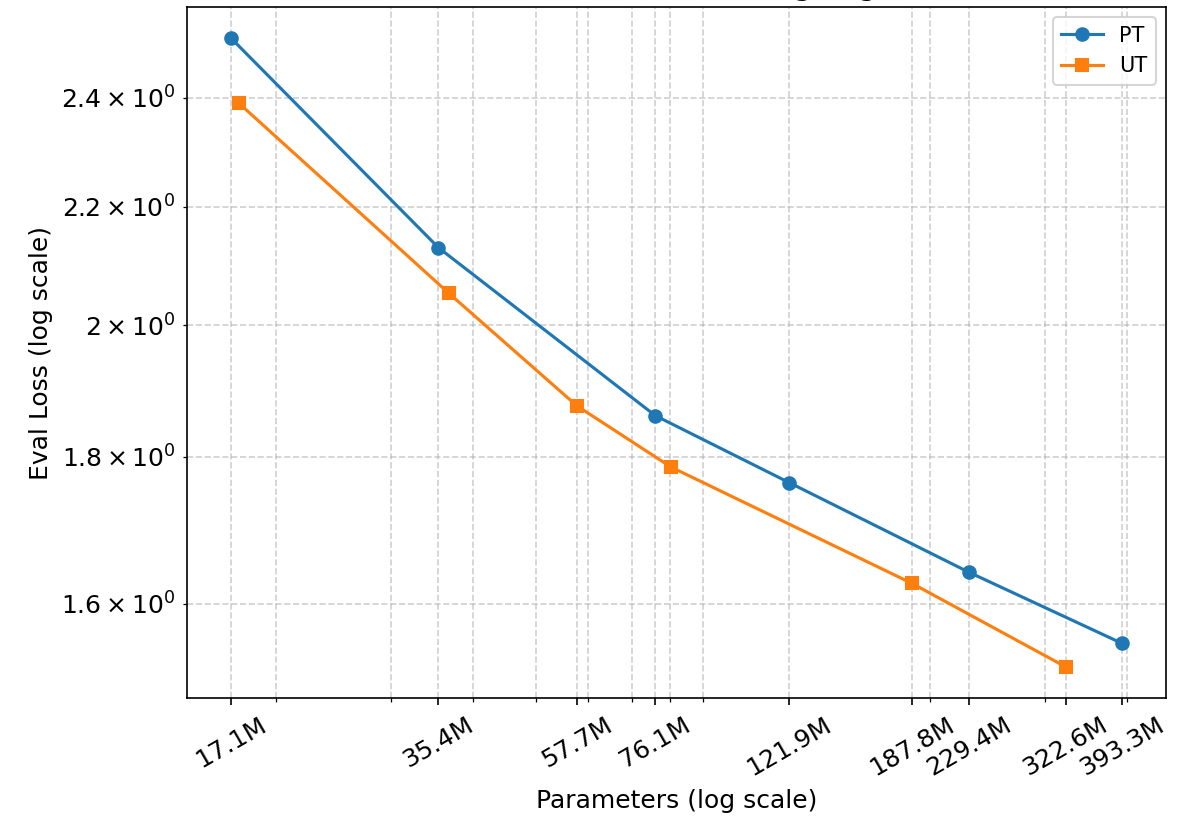}
        \caption{PT vs Universal Transformer.}
        \label{fig:right_plot2}
    \end{subfigure}
    
    \caption{
       Performance comparison between PT and BERT, and between PT and Universal Transformer across varying parameter scales. Both axes are on a log-log scale.
    }
    \label{fig:performance_comparison}
\end{figure*}

\subsection{Comparison with BERT and Universal Transformer}

\textbf{Tasks and Datasets.} We evaluated the performance of PT models across varying parameter scales on the Masked Language Modeling (MLM) task with a 15\% masking ratio. As comparative baselines, we trained BERT \cite{devlin2019bert} and Universal Transformer \cite{dehghani2019universal} models of equivalent parameter scales on the identical MLM task, utilizing the evaluation loss as the comparative metric. To ensure a fair comparison of data efficiency, all models were strictly limited to 1 training epoch on the MiniPile dataset \cite{kaddour2023minipile}.\\

\noindent
\textbf{Settings.} To validate the efficacy of our proposed cross-scale scaling framework, the PT architecture was scaled strictly adhering to the $\mu$P method. Consequently, we exclusively performed hyperparameter tuning on the smallest PT model ($\text{dim}_z=256$, approximately 17M parameters). Subsequently, this optimal hyperparameter configuration was zero-shot transferred to all larger-scale PT models. For BERT and the Universal Transformer, independent hyperparameter tuning was conducted at each parameter scale to guarantee optimal configurations across all dimensions.\\

\noindent
\textbf{Results.} Experimental results demonstrate that the performance of PT on the MLM task is consistently superior to that of the independently tuned BERT. This advantage is partially attributed to the architectural design of the PT: the iterative process of its variational inference inherently acts as cross-layer parameter sharing in a mathematical sense, enabling the PT to achieve higher computational volume (FLOPs) under an identical parameter count. However, compared to the Universal Transformer, which explicitly employs a recurrent cross-layer parameter sharing mechanism, the performance of the PT is relatively weaker. Detailed experimental comparative results are illustrated in Figure~\ref{fig:performance_comparison}.

\section{Related Works}

\textbf{Scaling and Hyperparameter Transfer} Modern large-scale models rely on Maximal Update Parametrization ($\mu$P) \cite{yang2022tensor} for zero-shot hyperparameter transfer. However, $\mu$P’s heuristic scaling often collapses the rigorous mathematical structures—such as probabilistic priors or energy constraints—inherent in white-box models. Developing analytical scaling rules that preserve these constrained architectures remains an underexplored challenge in optimization theory.\\

\noindent
\textbf{Mathematically Principled White-Box Scaling} Endowing neural networks with mathematical transparency often introduces optimization bottlenecks at scale. Implicit models like Deep Equilibrium Models (DEQs) \cite{bai2019deep,bai2020multiscale} and Continuous Hopfield Networks \cite{ramsauer2020hopfield} offer sound theoretical guarantees but suffer from hyperparameter sensitivity and instability when scaled, frequently requiring empirical regularizations that undermine their white-box nature.

Recently, CRATE-$\alpha$ \cite{yang2024scaling} successfully scaled white-box Transformers by introducing architectural modifications, such as overcomplete dictionaries and decoupled weights. While effective, this architecture-driven approach lacks a general analytical law for parameter optimization across scales. In contrast, our work focuses on the fundamental mathematical principles of parametrization. By reconstructing variational free energy and potential functions, we enable $\mu$P-level zero-shot transfer for Probabilistic Transformers (PT) without any architectural compromises. This provides a novel paradigm for scaling interpretable networks while maintaining their rigorous inference semantics.

\section{Conclusions}
\label{sec:bibtex}

We propose a modification to the architecture of PT to enable zero-shot hyperparameter transfer. This approach allows us to scale up PT at a significantly lower computational cost, even under limited resource constraints, and achieves superior performance on Masked Language Modeling (MLM) tasks compared to BERT \cite{devlin2019bert} at equivalent model scales.

\section*{Limitations}

A key architectural distinction between PT and the standard Transformer lies in PT’s iterative reuse of the same set of learnable parameters across multiple rounds—effectively implementing weight tying across layers, as opposed to the decoupled layer-wise parameters in the conventional Transformer. This design resembles that of the Universal Transformer \cite{dehghani2019universal}, which similarly employs parameter Sharing to increase computational depth without expanding the number of parameters. Consequently, under the same parameter budget, such tied-parameter models achieve greater effective computation and thus enhanced performance. As a result, while PT outperforms BERT \cite{devlin2019bert} on the MLM task, it still lags behind the Universal Transformer \cite{dehghani2019universal}.
Moreover, PT cannot leverage Flash Attention \cite{dao2022flashattention} during training, leading to significantly slower computation compared to standard Transformers.

\bibliography{references}

\begin{thebibliography}{15}
\providecommand{\natexlab}[1]{#1}

\bibitem[{Bai et~al.(2019)Bai, Kolter, and Koltun}]{bai2019deep}
S.~Bai, J.~Z. Kolter, and V.~Koltun. 2019.
\newblock Deep equilibrium models.
\newblock In \emph{Advances in Neural Information Processing Systems
  (NeurIPS)}, volume~32.

\bibitem[{Bai et~al.(2020)Bai, Koltun, and Kolter}]{bai2020multiscale}
S.~Bai, V.~Koltun, and J.~Z. Kolter. 2020.
\newblock Multiscale deep equilibrium models.
\newblock In \emph{Advances in Neural Information Processing Systems
  (NeurIPS)}, volume~33, pages 5238--15250.

\bibitem[{Bergstra and Bengio(2012)}]{bergstra2012random}
J.~Bergstra and Y.~Bengio. 2012.
\newblock Random search for hyper-parameter optimization.
\newblock \emph{Journal of Machine Learning Research (JMLR)}, 13(2).

\bibitem[{Blei et~al.(2017)Blei, Kucukelbir, and
  McAuliffe}]{blei2017variational}
David~M Blei, Alp Kucukelbir, and Jon~D McAuliffe. 2017.
\newblock Variational inference: A review for statisticians.
\newblock \emph{Journal of the American statistical Association},
  112(518):859--877.

\bibitem[{Blei et~al.(2003)Blei, Ng, and Jordan}]{blei2003latent}
David~M Blei, Andrew~Y Ng, and Michael~I Jordan. 2003.
\newblock Latent dirichlet allocation.
\newblock \emph{Journal of machine Learning research}, 3(Jan):993--1022.

\bibitem[{Dao et~al.(2022)Dao, Fu, Ermon, Rudra, and
  R{\'e}}]{dao2022flashattention}
Tri Dao, Daniel Fu, Stefano Ermon, Atri Rudra, and Christopher R{\'e}. 2022.
\newblock Flashattention: Fast and memory-efficient exact attention with
  io-awareness.
\newblock In \emph{Advances in Neural Information Processing Systems},
  volume~35, pages 10428--10440.

\bibitem[{Dehghani et~al.(2019)Dehghani, Gouws, Vinyals, Uszkoreit, and
  Kaiser}]{dehghani2019universal}
M.~Dehghani, S.~Gouws, O.~Vinyals, J.~Uszkoreit, and {\L}.~Kaiser. 2019.
\newblock Universal transformers.
\newblock In \emph{International Conference on Learning Representations
  (ICLR)}.

\bibitem[{Devlin et~al.(2019)Devlin, Chang, Lee, and
  Toutanova}]{devlin2019bert}
J.~Devlin, M.~W. Chang, K.~Lee, and K.~Toutanova. 2019.
\newblock {BERT}: Pre-training of deep bidirectional transformers for language
  understanding.
\newblock In \emph{Proceedings of the 2019 Conference of the North American
  Chapter of the Association for Computational Linguistics (NAACL-HLT)}.

\bibitem[{Kaddour et~al.(2023)Kaddour, Harris, Mozes, Bradley, Raileanu, and
  McHardy}]{kaddour2023minipile}
Jean Kaddour, Joshua Harris, Maximilian Mozes, Herbie Bradley, Roberta
  Raileanu, and Robert McHardy. 2023.
\newblock The minipile challenge for data-efficient language models.
\newblock \emph{arXiv preprint arXiv:2304.08442}.

\bibitem[{Lafferty et~al.(2001)Lafferty, McCallum, and
  Pereira}]{lafferty2001conditional}
John~D Lafferty, Andrew McCallum, and Fernando~CN Pereira. 2001.
\newblock Conditional random fields: Probabilistic models for segmenting and
  labeling sequence data.
\newblock In \emph{Proceedings of the Eighteenth International Conference on
  Machine Learning}.

\bibitem[{Ramsauer et~al.(2020)Ramsauer, Sch{\"a}fl, Lehner, Seidl, Widrich,
  Adler et~al.}]{ramsauer2020hopfield}
H.~Ramsauer, B.~Sch{\"a}fl, J.~Lehner, P.~Seidl, M.~Widrich, T.~Adler, and 1
  others. 2020.
\newblock Hopfield networks is all you need.
\newblock In \emph{International Conference on Learning Representations
  (ICLR)}.

\bibitem[{Vaswani et~al.(2017)Vaswani, Shazeer, Parmar, Uszkoreit, Jones,
  Gomez, Kaiser, and Polosukhin}]{vaswani2017attention}
Ashish Vaswani, Noam Shazeer, Niki Parmar, Jakob Uszkoreit, Llion Jones,
  Aidan~N Gomez, {\L}ukasz Kaiser, and Illia Polosukhin. 2017.
\newblock Attention is all you need.
\newblock In \emph{Advances in neural information processing systems},
  volume~30.

\bibitem[{Wu and Tu(2023)}]{wu2023probabilistic}
H.~Wu and K.~Tu. 2023.
\newblock Probabilistic transformer.
\newblock \emph{arXiv preprint arXiv:2305.10041}.

\bibitem[{Yang et~al.(2022)Yang, Hu, Babuschkin, Farhi, Ryder, Pachocki
  et~al.}]{yang2022tensor}
G.~Yang, E.~J. Hu, I.~Babuschkin, D.~Farhi, N.~Ryder, J.~Pachocki, and 1
  others. 2022.
\newblock Tensor programs v: Tuning large neural networks via zero-shot
  hyperparameter transfer.
\newblock In \emph{Advances in Neural Information Processing Systems
  (NeurIPS)}, volume~35, pages 13547--13585.

\bibitem[{Yang et~al.(2024)Yang, Li, Pai, Zhou, Ma, Yu, and
  Xie}]{yang2024scaling}
J.~Yang, X.~Li, D.~Pai, Y.~Zhou, Y.~Ma, Y.~Yu, and C.~Xie. 2024.
\newblock Scaling white-box transformers for vision.
\newblock In \emph{Advances in Neural Information Processing Systems
  (NeurIPS)}.

\end{thebibliography}

\appendix

\section{Equivalence Between Modified Variational Inference and Mathematical Essence}\label{app:A}
This section re-derives the Mean Field Variational Inference (MFVI) process to demonstrate that the reconstruction of potential functions and variational free energy mathematically aligns precisely with the core operation of the attention scaling factor in $\mu$P theory.

\subsection{Scaled Potential Functions and Joint Probability Distribution}

Under the $\mu$P framework of PT, we introduce a temperature parameter $\tau = N/r$. The redefined system potential functions are as follows:
\begin{enumerate}
    \item \textbf{Unary Factor}: $\phi_u(Z_i=a) = \exp(\tau S_{w_i, a})$
    \item \textbf{Ternary Factor}: $\phi_t(H_i=j, Z_i=a, Z_j=b) = \exp(\tau N T_{a,b}^{(c)})$ (if $H_i=j$, otherwise 1)
    \item \textbf{Binary Factor}: $\phi_b(Z_i=a, G_i=g) = \exp(\tau M B_{g,a})$
\end{enumerate}

Here, $T^{(c)}$ takes the low-rank decomposition form $U^{(c)}V^{(c)\top}$, $N$ is the hidden state dimension, $r$ is the rank of the ternary factor, and $M$ is the dimension of the binary factor. Given a sequence $w$, the joint probability distribution of the system $P(Z, H, G|w)$ is defined as:
\begin{equation}
\begin{split}
P(Z, H, G|w) = \frac{1}{\mathcal{Z}} \prod_{i=1}^n \phi_u(Z_i) \prod_{i=1}^n \phi_b(Z_i, G_i) \\
\times \prod_{c=1}^h \prod_{i=1}^n \prod_{j=1}^n \phi_t(H_i^{(c)}, Z_i, Z_j)
\end{split}
\end{equation}

\subsection{Reconstruction of Variational Free Energy Function}

To maintain the magnitude balance between the energy and entropy terms during large-scale expansion, we define the modified variational free energy $\mathcal{F}_{\mu P}(Q)$. For the marginal distribution $Q_i$ of the hidden variable $Z$, its free energy function incorporates the temperature coefficient $\tau$:
\begin{equation}
\mathcal{F}(Q_i) = \mathbb{E}_Q[-\ln P(Z, H, G|w)] - \tau H(Q_i)
\end{equation}
The fully expanded expected energy term $E$ is:
\begin{equation}
\begin{aligned} E = &-\sum_{i=1}^n \sum_{a=1}^N Q_i(a) \tau S_{w_i,a} \\ &-\sum_{c=1}^h \sum_{i=1}^n \sum_{j \neq i} Q_{ic}(j) \sum_{a=1}^N \sum_{b=1}^N Q_i(a) Q_j(b) (\tau N T_{a,b}^{(c)}) \\ &-\sum_{i=1}^n \sum_{a=1}^N \sum_{g=1}^M Q_i(a) Q_i^G(g) (\tau M B_{g,a}) \end{aligned}
\end{equation}

\subsection{Ternary Variable Update: Proof of Equivalence to Attention Scaling Factor $1/r$}

Consider the optimization of the ternary variable distribution $Q_{ic}(j)$ (i.e., attention weights). According to the MFVI iteration rules, its distribution is determined by the partial derivative of the energy term with respect to $Q_{ic}(j)$:
\begin{equation}
\frac{\partial E}{\partial Q_{ic}(j)} = -\sum_{a=1}^N \sum_{b=1}^N Q_i(a) Q_j(b) (\tau N T_{a,b}^{(c)})
\end{equation}
Substituting $\tau = N/r$ and the low-rank decomposition $T^{(c)} = \sum_{l=1}^r U_{a,l} V_{b,l}$:
\begin{equation}
\begin{split}
\frac{\partial E}{\partial Q_{ic}(j)} = -\frac{N^2}{r} \sum_{l=1}^r \left( \sum_{a=1}^N Q_i(a) U_{a,l} \right) \\
\times \left( \sum_{b=1}^N Q_j(b) V_{b,l} \right)
\end{split}
\end{equation}
To observe its scaling properties, we define the scaled quasi-distribution
\begin{equation}
    \tilde{Q}_i(a) = N Q_i(a) 
\end{equation}
since $Q_i(a)$ is a probability distribution with a coordinate magnitude of $\Theta(1/N)$, the coordinate magnitude of $\tilde{Q}_i(a)$ is restored to $\Theta(1)$. Let:
\begin{equation}
\begin{split}
\boldsymbol{q}_{i, \cdot} &= \sum_a \tilde{Q}_i(a) U_{a, \cdot} \\
\boldsymbol{k}_{j, \cdot} &= \sum_b \tilde{Q}_j(b) V_{b, \cdot}
\end{split}
\end{equation}
denote the Query and Key vectors, respectively. We then have:
\begin{equation}
\frac{\partial E}{\partial Q_{ic}(j)} = -\frac{1}{r} \boldsymbol{q}_i \boldsymbol{k}_j^\top
\end{equation}
Since $Q \propto \exp(-\nabla E)$, we obtain the update formulation:
\begin{equation}
Q_{ic}(j) \propto \exp \left( \frac{1}{r} \boldsymbol{q}_i \boldsymbol{k}_j^\top \right)
\end{equation}
In the $\mu$P implementation of the standard Transformer, the attention scores must be scaled by $qk^\top/d_k$. In the PT architecture, its "attention feature dimension" is determined by the rank $r$ of the ternary factor. The above derivation rigorously proves that: implicitly scaling the variational free energy is mathematically equivalent to setting the attention scaling factor to $1/r$ at the foundational computational graph level. This operation ensures that as $N \to \infty$, the attention scores do not cause Softmax saturation due to dot-product explosion.

\subsection{Hidden State Variable Update: Verification of Scale Stability}

For the distribution $Q_i(a)$ of the hidden variable $Z$, we construct the Lagrange function 
\begin{equation}
    L = \mathcal{F}(Q_i) + \lambda(\sum_a Q_i(a) - 1)
\end{equation}
Setting $\nabla_{Q_i(a)} L = 0$ yields:
\begin{equation}
\begin{split}
\nabla E_{Q_i(a)} - \tau (\ln Q_i(a) + 1) + \lambda = 0 \\
\Rightarrow \quad Q_i(a) \propto \exp \left( -\frac{1}{\tau} \nabla E_{Q_i(a)} \right)
\end{split}
\end{equation}
where each term in the energy gradient $\nabla E_{Q_i(a)}$ includes the coefficient $\tau$. Upon expansion, the Logits for $Q_i(a)$ are:
\begin{equation}
\begin{split}
\text{Logits}_i(a) &= \frac{1}{\tau} \Big[ \tau S_{w_i,a}+ \tau M \sum_{g=1}^M Q_i^G(g) B_{g,a} \\
&\quad + \tau N \sum_{c,j} Q_{ic}(j) \sum_{b=1}^N Q_j(b) T_{a,b}^{(c)} \Big] \\
&= S_{w_i,a} + \sum_{g=1}^M (M Q_i^G(g)) B_{g,a} \\
&\quad + \sum_{c,j} Q_{ic}(j) \sum_{b=1}^N (N Q_j(b)) T_{a,b}^{(c)}
\end{split}
\end{equation}
It is evident that the temperature parameter $\tau$ cancels out between the numerator and denominator. This implies:
\begin{enumerate}
    \item \textbf{Dimension-Adaptive Quasi-Distribution Scaling}: The binary and ternary terms in the update equation automatically carry their corresponding dimension scaling factors (i.e., $\tilde{Q}_z = N Q_z$ and $\tilde{Q}_G = M Q_G$), precisely converting probabilities into feature vectors with a constant coordinate intensity of $\Theta(1)$.
    \item \textbf{Absolute Scale Invariance}: The magnitude of the Logits is entirely determined by the parameter matrices (e.g., $S, T, B$) and is fully decoupled from the model widths $N$ and $M$.
\end{enumerate}

\section{Unified Adaptability of Potential Function and Variational Free Energy Modifications in the Head Selection Module}
\label{app:B}

In the Head Selection module of PT, due to the expansion of the model width $N$, two different parametrization scaling paradigms can be adopted: the first is scaling the channel count $C$ ($C \propto N, r = \Theta(1)$), and the second is scaling the rank $r$ of the ternary factor ($C = \Theta(1), r \propto N$). This section details the derivation and proves that by introducing the temperature parameter $\tau = N/r$, the modified variational free energy ensures a magnitude balance between the energy and entropy functions throughout the entire training cycle under both paradigms.

The variational free energy function is defined as:
\begin{equation}
\mathcal{F}(Q) = E - \tau H(Q)
\end{equation}
where the expected energy $E$ expands into the sum of unary, binary, and ternary potential terms:
\begin{equation}
\begin{split}
E = &-\sum_i \sum_{a=1}^N Q_i(a) \tau S_{w_i,a} \\
    &-\sum_{i} \sum_{a,g} Q_i(a) Q_i^G(g) \tau M B_{g,a} \\
    &-\sum_{c=1}^C \sum_{i,j} Q_{ic}(j) \sum_{a,b} Q_i(a) Q_j(b) \tau N T_{a,b}^{(c)}
\end{split}
\end{equation}
To facilitate magnitude analysis, we restore the probability distributions to quasi-distributions of scale $\Theta(1)$ by setting $\tilde{Q}_i(a) = N Q_i(a)$ and $\tilde{Q}_i^G(g) = M Q_i^G(g)$.

\subsection{Detailed Derivation of Paradigm 1: Scaling Channel Count $C$}

In this paradigm, $C \propto N$ and $r = \Theta(1)$. Consequently, the temperature parameter is $\tau = \Theta(N)$.

\textbf{1. Magnitude of the Entropy Function}

For the entropy function 
\begin{equation}
    H(Q) = -\sum_{a=1}^N Q(a) \ln Q(a)
\end{equation}
we consider its evolution in stages.

\textbf{Early Training Stage (Uniform Distribution)} At this point, 
\begin{equation}
    H(Q) = \sum_{a=1}^N \frac{1}{N} \ln N = \ln N
\end{equation}
Combined with the setting $\tau = \Theta(N)$, its overall magnitude is
\begin{equation}
    \tau H(Q) = \Theta(N \ln N)
\end{equation}

\textbf{Convergence Stage (Confident Distribution)} At this point, the distribution tends to be sparse, with a few labels occupying the vast majority of the probability, yielding $H(Q) = \Theta(1)$. Correspondingly, the overall magnitude becomes 
\begin{equation}
    \tau H(Q) = \Theta(N)
\end{equation}

\textbf{2. Magnitude of the Unary Potential Term}

For the unary potential term 
\begin{equation}
    E_{unary} = -\frac{\tau}{N} \sum_{a=1}^N \tilde{Q}_i(a) S_{w_i,a}
\end{equation}
the derivation is as follows.

\textbf{Early Training Stage} The $N$ individual terms of parameter $S$ can be viewed as mutually independent, with their sum exhibiting a random walk behavior. The variance is 
\begin{equation}
    \mathrm{Var}(\sum \tilde{Q} S) = N \cdot \mathrm{Var}(\tilde{Q} S) = \Theta(N)
\end{equation}
making the sum of absolute values $\Theta(\sqrt{N})$. Substituting the coefficients gives 
\begin{equation}
    E_{unary} = \frac{\Theta(N)}{N} \cdot\Theta(\sqrt{N}) = \Theta(\sqrt{N})
\end{equation}

\textbf{Convergence Stage} The variables $\tilde{Q}$ and $S$ tend to align after gradient updates, and their corresponding signs become highly correlated (coherent superposition). Therefore, 
\begin{equation}
    \sum_{a=1}^N \tilde{Q}_i(a) S_{w_i,a} = \sum_{a=1}^N \Theta(1) = \Theta(N)
\end{equation}
Substituting the coefficients yields the final magnitude
\begin{equation}
    E_{unary} = \frac{\Theta(N)}{N}\cdot \Theta(N) = \Theta(N)
\end{equation}

\textbf{3. Magnitude of the Binary Potential Term}

For the binary potential term 
\begin{equation}
    E_{binary} = -\frac{\tau}{N} \sum_{a=1}^N \tilde{Q}_i(a) \sum_{g=1}^M \tilde{Q}_i^G(g) B_{g,a}
\end{equation}
the derivation is as follows.

\textbf{Early Training Stage} According to the $\mu$P initialization principles, 
\begin{equation}
    B_{g,a} = \Theta(1/\sqrt{N})
\end{equation}
Independent random walks lead the inner sum to $\Theta(\sqrt{M/N}) = \Theta(1)$ (since $M \propto N$), and the random walk of the total sum yields $\Theta(\sqrt{N})$. After substituting the coefficients, 
\begin{equation}
    E_{binary} = \Theta(\sqrt{N})
\end{equation}

\textbf{Convergence Stage} The parameter is decomposed as
\begin{equation}
    B = B^{init} + B^{learn}
\end{equation}
where $B^{learn} = \Theta(1/N)$. The inner sum $\sum_g \tilde{Q}_i^G(g) B_{g,a}^{learn}$ generates an aligned accumulation of 
\begin{equation}
    \sum_g \Theta(1)\Theta(1/N) = \Theta(1)
\end{equation}
The outer sum undergoes another coherent superposition, producing $\Theta(N)$. The final magnitude is 
\begin{equation}
    E_{binary} = \frac{\Theta(N)}{N}\cdot \Theta(N) = \Theta(N)
\end{equation}

\textbf{4. Magnitude of the Ternary Potential Term}

For the single-channel energy
\begin{equation}
    E_{ternary}^{(c)} = -\frac{\tau}{N} \sum_{i,j} Q_{ic}(j) \sum_{a,b} \tilde{Q}_i(a) \tilde{Q}_j(b) T_{a,b}^{(c)}
\end{equation}

the derivation is as follows.

\textbf{Convergence Stage} Utilizing the low-rank decomposition
\begin{equation}
    T_{a,b} = \sum_{l=1}^r U_{a,l} V_{b,l}
\end{equation}
Because $r = \Theta(1)$, the aligned learned parts are $U^{learn}, V^{learn} = \Theta(1/N)$. After the double superposition computation of $\sum_{a,b}$, the sum of the inner dot products attains a magnitude of $\Theta(1)$. Since $\tau/N = \Theta(1)$, the single-channel energy is $\Theta(1)$. With multi-channel accumulation $C \propto N$, the total ternary energy is 
\begin{equation}
    E_{ternary} = C \cdot \Theta(1) = \Theta(N)
\end{equation}

\textbf{Conclusion for Paradigm 1} In the early training stage, the entropy term $\Theta(N \ln N)$ is larger in magnitude than the energy term $\Theta(\sqrt{N})$, thereby driving the distribution away from the uniform state. During the mid-to-late training stages, the energy and entropy terms become strictly equal, both reaching $\Theta(N)$, which maintains a healthy distribution and prevents the severe magnitude imbalance inevitably caused when $\tau=1$.

\subsection{Detailed Derivation of Paradigm 2: Scaling Rank $r$}

In this scheme, $C = \Theta(1)$ and $r \propto N$. Thus, the temperature parameter is 
$\tau = \frac{N}{r} = \Theta(1)$. The detailed derivation process is analogous to Paradigm 1.

\textbf{1. Magnitude of the Entropy Function}

Based on the setting $\tau = \Theta(1)$:

\textbf{Early Training Stage} The overall magnitude of the entropy term is 
\begin{equation}
    \tau H(Q) = \Theta(1) \cdot \ln N = \Theta(\ln N)
\end{equation}

\textbf{Convergence Stage} The overall magnitude of the entropy term is 
\begin{equation}
    \tau H(Q) = \Theta(1) \cdot \Theta(1) = \Theta(1)
\end{equation}

\textbf{2. Magnitude of the Energy Function (Unary and Binary Potentials)}

\textbf{Early Training Stage} Following the derivation in Paradigm 1, the unscaled original sum of energies is $\Theta(\sqrt{N})$. Substituting the new coefficient 
\begin{equation}
    \frac{\tau}{N} = \frac{\Theta(1)}{N}
\end{equation}
we obtain
\begin{equation}
    E_{unary, binary} = \Theta(\frac{1}{\sqrt{N}}) \approx 0
\end{equation}

\textbf{Convergence Stage} The original aligned sum of energies is $\Theta(N)$. Substituting the new coefficient yields
\begin{equation}
    E_{unary, binary} = \frac{\Theta(1)}{N} \cdot \Theta(N) = \Theta(1)
\end{equation}

\textbf{3. Magnitude of the Energy Function (Ternary Potential)}

\textbf{Convergence Stage} Since $r \propto N$, the learned part of the transition matrix is
\begin{equation}
    T_{a,b}^{learn} = \sum_{l=1}^r U_{a,l} V_{b,l} \sim r \cdot (\frac{1}{N} \cdot \frac{1}{N}) = \Theta(\frac{1}{N})
\end{equation}
The double summation $\sum_{a,b=1}^N \tilde{Q}(a)\tilde{Q}(b) T_{a,b}$ is equivalent to the coherent superposition of $N^2$ terms, yielding an original magnitude of
\begin{equation}
    N^2 \cdot \Theta(\frac{1}{N}) = \Theta(N)
\end{equation}
Applying the ternary term coefficient $\frac{\tau}{N} = \frac{1}{r}$ (this operation corresponds to the underlying computational graph's Attention $1/d_k$ scaling), the single-channel energy becomes 
\begin{equation}
    \frac{1}{r} \cdot \Theta(N) = \Theta(1)
\end{equation}
Because the channel count $C = \Theta(1)$, the total ternary energy is $\Theta(1)$.

\textbf{Conclusion for Paradigm 2} In the early training stage, the entropy term $\Theta(\ln N)$ is significantly larger than the energy term (which approaches 0), propelling the distribution towards rapid convergence. During the mid-to-late training stages, both terms stabilize at the $\Theta(1)$ magnitude.

\section{Theoretical Proof of PT Satisfying $\mu$P Principles}\label{app:C}
To ensure that the Probabilistic Transformer (PT) possesses cross-scale transferability of optimal hyperparameters during width $N$ expansion, we adhere to the guiding principles of Maximal Update Parametrization ($\mu$P). This section provides rigorous mathematical derivations to prove how PT's parameterization adjustment scheme perfectly satisfies the three fundamental principles of $\mu$P from both initialization and dynamic perspectives.

\subsection{Parameter Classification, Initialization, and Learning Rate Strategies}

We categorize all learnable parameters of PT into the following three groups based on their positions in the computational graph, tensor shapes, and impacts on activation magnitudes. We assign distinct initialization standard deviations $\sigma$ and learning rate scaling factors $\eta_{\text{mult}}$ to each group:

\begin{enumerate}
    \item \textbf{Input Group}: Primarily includes the unary potential matrix $S \in \mathbb{R}^{V \times N}$ and all biases. These parameters map discrete or low-dimensional inputs to a high-dimensional space.
    \begin{itemize}
        \item \textbf{Initialization}: Set standard deviation $\sigma_{in} = \Theta(1)$.
        \item \textbf{Learning Rate}: Set scaling factor $\eta_{in\_mult} = 1$.
    \end{itemize}

    \item \textbf{Hidden Group}: Includes the low-rank decomposition matrices $U, V \in \mathbb{R}^{N \times r}$ of the ternary factor and the binary factor matrix $B \in \mathbb{R}^{M \times N}$ (where $M \propto N$). These parameters dictate interactions between high-dimensional representations.
    \begin{itemize}
        \item \textbf{Initialization}: Set standard deviation $\sigma_{hid} = \Theta(1/\sqrt{N})$.
        \item \textbf{Learning Rate}: Set scaling factor $\eta_{hid\_mult} = 1/N$.
    \end{itemize}

    \item \textbf{Output Group}: Refers to the decoding matrix $W_{out} \in \mathbb{R}^{N \times V}$ in the prediction head. These parameters project high-dimensional representations back to scalar scores.
    \begin{itemize}
        \item \textbf{Initialization}: Set standard deviation $\sigma_{out} = \Theta(1/N)$.
        \item \textbf{Learning Rate}: Set scaling factor $\eta_{\text{out\_mult}} = 1$.
    \end{itemize}
\end{enumerate}

\subsection{Principle 1: Coordinate Magnitude Stability (Forward Pass Stability)}

Principle 1 Requirement: At initialization, the magnitude of the activation (coordinate) of any neuron in the neural network must remain at $\Theta(1)$.

During the PT iteration process, consider the linear combination term of the hidden layer neurons 
\begin{equation}
    y_i = \sum_{j=1}^N W_{ij} x_j
\end{equation}
such as message passing in ternary factors. Assuming the input coordinates $x_j$ satisfy $\Theta(1)$, based on the properties of the sum of independent random variables, its output variance satisfies:
\begin{equation}
\text{Var}(y_i) = \sum_{j=1}^N \text{Var}(W_{ij}) \mathbb{E}[x_j^2] = N \cdot \sigma_{hid}^2 \cdot \Theta(1)
\end{equation}
Substituting the hidden layer initialization strategy $\sigma_{hid} = \Theta(1/\sqrt{N})$ defined in C.1, we get:
\begin{equation}
\text{Var}(y_i) = N \cdot \frac{1}{N} \cdot \Theta(1) = \Theta(1)
\end{equation}
This proves that when PT expands in width, the coordinate intensity of its internal activations remains stable at $\Theta(1)$, ensuring the numerical stability of the forward pass.

\subsection{Principle 2: Boundedness of Output Scores (Output Score Stability)}

Principle 2 Requirement: The magnitude of the model's final output Logits should be $O(1)$ and must not explode as width increases.

In the output layer, the predicted score is 
\begin{equation}
    s = \sum_{j=1}^N W_{out, j} y_j
\end{equation}
Given that the activations $y_j \sim \Theta(1)$, the variance of the output score is:
\begin{equation}
\text{Var}(s) = N \cdot \sigma_{out}^2 \cdot \text{Var}(y_j) = N \cdot \frac{1}{N^2} \cdot \Theta(1) = \Theta(1/N)
\end{equation}
Therefore, at initialization, the magnitude of the output scores is $O(1/\sqrt{N})$. As training progresses, this magnitude will evolve to $\Theta(1)$, but since it is constrained by the upper bound of $O(1)$, the Softmax layer will not suffer from saturation or vanishing gradients due to the expansion of width $N$.

\subsection{Principle 3: Update Magnitude Stability Based on AdamW (Update Stability)}

Principle 3 Requirement: After one round of parameter updating, the change in activations
\begin{equation}
    \Delta y = \Delta W \cdot x
\end{equation}
must remain at $\Theta(1)$.

For the \textbf{AdamW optimizer}, the magnitude of the single-coordinate update $\Delta W_{ij}$ for weights predominantly depends on the learning rate $\eta$. During the feature learning phase, there is a significant coherence between gradients and input activations.

\begin{enumerate}
    \item \textbf{Hidden Layer Parameter Updates}: For the hidden layer, the change in its activations is:
    \begin{equation}
    \Delta y_i = \sum_{j=1}^N \Delta W_{ij} x_j \approx \eta_{hid} \sum_{j=1}^N \text{sign}(g_{ij}) x_j
    \end{equation}
    Due to the coherent superposition effect, the magnitude of the summation term reaches $\Theta(N)$. To achieve $\Delta y_i = \Theta(1)$, the following condition must be met:
    \begin{equation}
    \eta_{hid} = \eta_{base} \cdot \eta_{hid\_mult} = \eta_{base} \cdot \frac{1}{N}
    \end{equation}
    This explains why the learning rate of the hidden layers must scale inversely proportional to the width.

    \item \textbf{Input and Output Layer Updates}: The updates to the input layer (unary potential terms) do not involve linear accumulation across dimensions; their magnitude of change is directly determined by $\eta_{in}$. Thus, keeping $\eta_{in\_mult} = 1$ is sufficient to achieve an update of $\Theta(1)$. For the output layer, to counteract the $1/N$ suppression at initialization and ensure it can quickly learn effective logic scores, we set $\eta_{out\_mult} = 1$. This allows the feature weights of the output layer to grow rapidly in the early stages of training and stabilize at $\Theta(1)$.
\end{enumerate}

\end{document}